\documentclass[10pt,twocolumn,letterpaper]{article}

\usepackage{arxiv}
\usepackage{times}
\usepackage{epsfig}
\usepackage{graphicx}
\usepackage{amsmath}
\usepackage{amssymb}

\usepackage{comment}
\usepackage[british,UKenglish,USenglish,english,american]{babel}
\usepackage{multirow}

\usepackage{makecell}
\usepackage{subfigure}
\usepackage{bm}

\usepackage[pagebackref=true,breaklinks=true,letterpaper=true,colorlinks,bookmarks=false]{hyperref}

\begin{document}

\title{An Empirical Study of Spatial Attention Mechanisms in Deep Networks}

\author{Xizhou Zhu$^{1,3\dag}$\thanks{Equal contribution. \dag This work is done when Xizhou Zhu and Dazhi Cheng are interns at Microsoft Research Asia.}\quad Dazhi Cheng$^{2,3\dag*}$ \quad Zheng Zhang$^{3*}$ \quad Stephen Lin$^3$ \quad Jifeng Dai$^3$ \vspace{8pt}\\
	$^1$University of Science and Technology of China\\
	$^2$Beijing Institute of Technology\\
	$^3$Microsoft Research Asia\\
	{\tt\small ezra0408@mail.ustc.edu.cn} \\
	{\tt\small \{v-dachen,zhez,stevelin,jifdai\}@microsoft.com} \\	
}

\maketitle

\begin{abstract}
Attention mechanisms have become a popular component in deep neural networks, yet there has been little examination of how different influencing factors and methods for computing attention from these factors affect performance. Toward a better general understanding of attention mechanisms, we present an empirical study that ablates various spatial attention elements within a generalized attention formulation, encompassing the dominant Transformer attention as well as the prevalent deformable convolution and dynamic convolution modules. Conducted on a variety of applications, the study yields significant findings about spatial attention in deep networks, some of which run counter to conventional understanding. 
For example, we find that the comparison of query and key content in Transformer attention is negligible for self-attention, but vital for encoder-decoder attention. On the other hand, a proper combination of deformable convolution with key content saliency achieves the best accuracy-efficiency tradeoff in self-attention. Our results suggest that there exists much room for improvement in the design of attention mechanisms.
\end{abstract}

\section{Introduction}

\begin{figure*}
\begin{center}
\includegraphics[width=0.95\linewidth]{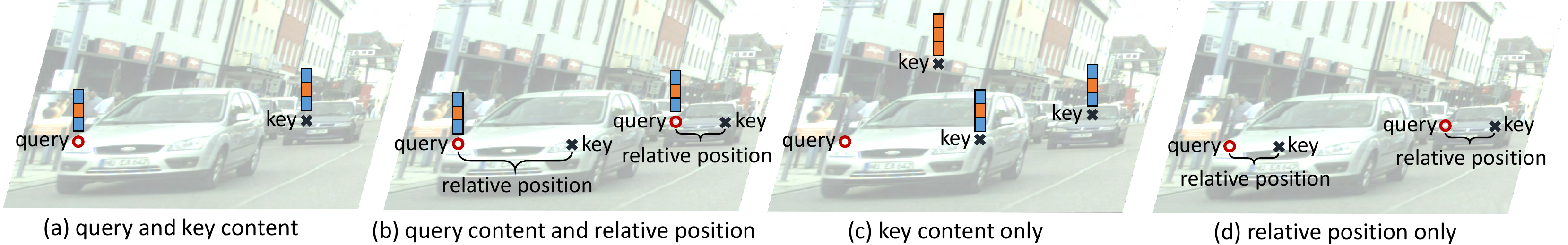}
\end{center}
\vspace{-0.5em}
\caption{Illustration of different attention terms. The color bar above a sampling point denotes its content feature. The existence of content features and/or relative position indicates that the term uses them for attention weight calculation.}
\vspace{-0.8em}
\label{fig.attention_illustration}
\end{figure*}

Attention mechanisms enable a neural network to focus more on relevant elements of the input than on irrelevant parts. They were first studied in natural language processing (NLP), where encoder-decoder attention modules were developed to facilitate neural machine translation~\cite{bahdanau2015neural,luong2015effective,gehring2017convolutional}. In computing the output for a given query element (\eg, a target word in the output sentence), certain key elements (\eg, source words in the input sentence) are prioritized according to the query. Later, self-attention modules were presented for modeling intra-sentence relations~\cite{cheng2016long,lin2017structured,parikh2016decomposable,paulus2017deep,vaswani2017attention}, where both the key and query are from the same set of elements. In a milestone paper~\cite{vaswani2017attention}, the Transformer attention module is presented, superseding past works and substantially surpassing their performance. The success of attention modeling in NLP has led to its adoption in computer vision, where different variants of Transformer attention are applied to recognition tasks such as object detection and semantic segmentation~\cite{hu2018relation,wang2018non,gu2018learning,huang2018ccnet,zhao2018psanet,fu2018dual}, where the query and key are visual elements such as image pixels or regions of interest.

In determining the attention weight assigned to a certain key for a given query, there exist just a few properties of the input that are commonly considered. One is the content of the query. For the case of self-attention, the query content may be the features at the query pixel in an image, or of a word in a sentence. Another is the content of the key, where a key may be a pixel within a local neighborhood of the query, or another word within the sentence. The third is the relative position of the query and key.

Based on these input properties, there are four possible {\em attention factors} from which the attention weight for a key with respect to a query is determined, as these factors must account for information about the key. Specifically, these factors are (1) the query and key content, (2) the query content and relative position, (3) the key content only, and (4) the relative position only. In the latest version of Transformer attention~\cite{dai2019transformer}, attention weights are expressed as a sum of four terms ($\mathcal{E}_{1}$, $\mathcal{E}_{2}$, $\mathcal{E}_{3}$, $\mathcal{E}_{4}$), one for each of these attention factors as illustrated in Fig.~\ref{fig.attention_illustration}. The nature of the dependencies involved with these terms vary. For example, the first two ($\mathcal{E}_{1}$, $\mathcal{E}_{2}$) are sensitive to the query content. While, the latter two ($\mathcal{E}_{3}$, $\mathcal{E}_{4}$) do not account for query content, but rather they mainly capture salient key elements and exploit global positional biases, respectively. Although attention weights can be decomposed into terms based on these factors, their relative significance in various inference problems has not been closely examined in the literature. Moreover, prevalent modules like deformable convolution~\cite{dai2017deformable,zhu2018deformable} and dynamic convolution~\cite{wu2019pay}, though seemingly orthogonal to Transformer attention,
also employ mechanisms that focus on certain parts of an input.
Whether these modules can all be viewed from a unified perspective and how their operational mechanisms differ also have not been explored. 

In this work, we perceive Transformer attention, deformable convolution, and dynamic convolution modules as various instantiations of spatial attention, involving different subsets of the attention factors and accounting for these factors in different ways. Towards disentangling the effects of different attention factors and mechanisms, we present an empirical study of spatial attention, in which various elements of attention mechanisms are ablated within a generalized attention formulation. This investigation is conducted on a variety of applications, namely neural machine translation, semantic segmentation, and object detection. From this study, we find that: 1) In the Transformer attention module, the query-sensitive terms, especially the query and key content term, play a minor role in self-attention. But in encoder-decoder attention, the query and key content term is vital. 2) Though deformable convolution utilizes an attention mechanism based only on the query content and relative position term, it operates more effectively and efficiently on image recognition than the counterpart in Transformer attention. 3) In self-attention, the factors of query content \& relative position and key content only are the most important. A proper combination of deformable convolution and the key content only term in Transformer attention delivers higher accuracy than that of the Transformer attention module, with much lower computational overhead on image recognition tasks.

The observations made in this paper challenge the conventional understanding of current spatial attention mechanisms. For example, it is widely believed that their success can mainly be attributed to query-sensitive attention, especially the query and key content term. This understanding perhaps originates from the initial success of the encoder-decoder attention module in neural machine translation. Thus, in some recent variants~\cite{wang2018non,huang2018ccnet,yuan2018ocnet,fu2018dual}, like the non-local block~\cite{wang2018non} and criss-cross attention module~\cite{huang2018ccnet}, only the query and key content term is kept, with all the other terms removed. These modules still function well in self-attention applications, which strengthen this perception. However, our study suggests that this understanding is incorrect.  We find that these attention modules with only query-sensitive terms actually perform on par with those with only query-irrelevant terms. Our study further suggests that this degeneration is likely due to the design of the attention modules, rather than an inherent characteristic of self-attention, since deformable convolution is found to exploit query content \& relative position effectively and efficiently in image recognition tasks.

This empirical analysis suggests that there is much room for improvement in the design of spatial attention mechanisms in deep networks. Its findings are used in this paper to make some initial headway in this direction, and it is hoped that this study will spur further investigation into the operational mechanisms used in modeling spatial attention.

\section{Related Work}

\noindent\textbf{Development and application of attention-based modules.}
The field of NLP has witnessed steady development of attention mechanisms in recent years~\cite{bahdanau2015neural,luong2015effective,gehring2017convolutional,vaswani2017attention,shaw2018self,dai2019transformer}. Starting from the introduction of an attention module in neural machine translation~\cite{bahdanau2015neural}, various attention factors and weight assignment functions based on these factors have been utilized. In~\cite{luong2015effective}, the inner product of vectors encoding query and key contents is recommended for computing attention weights, and absolute spatial positions are incorporated as an attention factor. In~\cite{gehring2017convolutional}, the weight assignment additionally accounts for the inner product of spatial positions encoded in high-dimensional vectors. The landmark work of Transformer~\cite{vaswani2017attention} set a new standard, and its latest variants use relative positions instead of absolute positions for better generalization ability~\cite{shaw2018self,dai2019transformer}. In this paper, we conduct the empirical study on the latest instantiation of Transformer attention~\cite{dai2019transformer} from this family of works.

Motivated by their success in NLP tasks~\cite{bahdanau2015neural,luong2015effective,gehring2017convolutional,cheng2016long,lin2017structured,parikh2016decomposable,paulus2017deep,vaswani2017attention}, attention mechanisms have also been employed in computer vision applications such as relational reasoning among objects~\cite{battaglia2016interaction,santoro2017simple}, image captioning~\cite{xu2015show}, image generation~\cite{zhang2018self,xu2018attngan}, image recognition~\cite{hu2018relation,wang2018non,gu2018learning,huang2018ccnet,zhao2018psanet,fu2018dual}, and video recognition~\cite{zhu2017flow,xiao2018video}. In vision, the key and query refer to visual elements, but aside from that, most of these works use a formulation similar to Transformer attention. Since the effects of different attention module elements may vary with the target application, we conduct the empirical study on three different tasks that have been influenced greatly by attention modeling, namely neural machine translation in NLP, and object detection and semantic segmentation in computer vision.

Aside from Transformer attention, there are variants of convolution, such as deformable convolution~\cite{dai2017deformable,zhu2018deformable} and dynamic convolution~\cite{wu2019pay}, that also can be viewed as types of attention mechanisms which operate on a subset of the attention factors using different attention weight functions. They also are included in the study for examination.

It is worth mentioning a dual form of spatial attention, called channel-wise feature attention~\cite{wang2017residual,zhang2018context,hu2018squeeze,fu2018dual}. As different feature channels encode different semantic concepts, these works seek to capture the correlations among these concepts through activation/deactivation of certain channels. Meanwhile, in the spatial domain, relationships among elements at different spatial positions are modeled, with the same attention weights on feature channels assigned to related spatial positions. The development of channel-wise feature attention has been focused on certain image recognition tasks, like semantic segmentation and image classification. In this paper, our empirical study specifically examines spatial attention mechanisms designed for broad application.

\vspace{0.5em}
\noindent\textbf{Analysis of spatial attention mechanisms.} There exists relatively little analysis of spatial attention mechanisms despite their prevalence in deep networks. This research has largely been conducted by visualizing or analyzing the learned attention weights of a whole attention module on only NLP tasks~\cite{ghader2017does,tang2018analysis,ghaeini2018interpreting,jain2019attention}. Many works~\cite{ghader2017does,tang2018analysis,ghaeini2018interpreting} suggest that attention weight assignment in encoder-decoder attention plays a role similar to word alignment in traditional approaches~\cite{alkhouli2016alignment,cohn2016incorporating,liu2016neural,chen2016guided}. The implicit underlying assumption in these works is that the input elements accorded high attention weights are responsible for the model outputs. However, recent research casts doubt on this assumption~\cite{jain2019attention}, finding that attention weights do not correlate well with feature importance measures, and that counterfactual attention weight configurations do not yield corresponding changes in prediction.

In this paper, we conduct the first comprehensive empirical study on the elements of spatial attention modules over both NLP and computer vision tasks. Different attention factors and weight assignment functions are carefully disentangled, with their effects directly measured by the final performance on these tasks. 

\section{Study of Spatial Attention Mechanisms}

To facilitate our study, we develop a generalized attention formulation that is able to represent various module designs. We then show how the dominant attention mechanisms can be represented within this formulation, and how ablations can be conducted using this formulation with respect to different attention module elements.

\vspace{0.5em}
\noindent\textbf{Generalized attention formulation}

Given a query element and a set of key elements, an attention function adaptively aggregates the key contents according to attention weights that measure the compatibility of query-key pairs. To allow the model to attend to key contents from different representation subspaces and different positions, the outputs of multiple attention functions (heads) are linearly aggregated with learnable weights. Let $q$ index a query element with content $z_q$, and $k$ index a key element with content $x_k$. Then the multi-head attention feature $y_q$ is computed as 
\begin{equation}
\small
y_q = \sum_{m=1}^{M}W_m \big[ \sum_{k\in \Omega_q}A_m(q,k,z_q,x_k) \odot W_m' x_k \big],
\label{eq:general_attention}
\end{equation}
where $m$ indexes the attention head, $\Omega_q$ specifies the supporting key region for the query, $A_m(q,k,z_q,x_k)$ denotes the attention weights in the $m$-th attention head, and $W_m$ and $W_m'$ are learnable weights. Usually, the attention weights are normalized within $\Omega_q$, as $\sum_{k\in \Omega_q}A_m(q,k,z_q,x_k) = 1$.

In encoder-decoder attention, the key and the query are from two different sets of elements, where in most applications the two sets of elements need to be properly aligned. For example, in the encoder-decoder attention of neural machine translation, the key and the query elements correspond to the words in the input and the output sentences, respectively, where proper alignment is necessary for correct translation. Meanwhile, in self-attention, the key and the query are from the same set of elements. For example, both the key and the query are of words in the input or output sentence. In such scenarios, the self-attention mechanism is expected to capture intra-relationships among the elements, and usually the query and the key contents are modeled by the same set of features, \ie, $x=z$.

\setlength{\tabcolsep}{3pt}
\renewcommand{\arraystretch}{1.0}
\begin{table*}[t]
        \centering
        \small
        \resizebox{0.9\linewidth}{!}{
        \begin{tabular}{c|c|c|c|c|c|c}
        \Xhline{2\arrayrulewidth}
        \multicolumn{2}{c|}{attention mechanism} & spatial properties & query content &  key content & relative position & complexity\\ 
                \hline  
                \multirow{4}{*}{Transformer attention} & $\mathcal{E}_{1}$ & dense, global & \checkmark & \checkmark &  &  $O(N_s^2C + N_sC^2)$\\  
                \cline{2-7}
                                                    & $\mathcal{E}_{2}$ & dense, global & \checkmark &  & \checkmark & $O(N_s^2C + N_sC^2)$\\  
                \cline{2-7}
                                                    & $\mathcal{E}_{3}$ & dense, global &  & \checkmark &  &  $O(N_sC^2)$\\  
                \cline{2-7}
                                                    & $\mathcal{E}_{4}$ & dense, global &  &  & \checkmark & $O(N_s^2C + N_sC^2)$\\  
                \hline
                \multicolumn{2}{c|}{Regular convolution} & sparse, local &   &  & \checkmark & $O(N_sC^2N_k)$\\  
                \hline
                \multicolumn{2}{c|}{Deformable convolution} & sparse, global & \checkmark &  & \checkmark &  $O(N_sC^2N_k)$\\  
                \hline
                \multicolumn{2}{c|}{Dynamic convolution} & sparse, local & \checkmark &  & \checkmark & $O(N_sCN_gN_k + N_sC^2)$\\  
        \Xhline{2\arrayrulewidth}
        \end{tabular}
        }
        \vspace{.5em}
        \caption{Comparison of different attention mechanisms. $N_s$ denotes number of spatial elements, i.e. width by height for images, and number of tokens for text; $C$ denotes representation dimension; $N_k$ denotes kernel size of convolution ($N_k=3 \times 3$ for images and $N_k=3$ for text, by default); $N_g$ denotes number of feature groups in dynamic convolution.}
        \label{table:compare_attentions}
        \vspace{-1.5em}
\end{table*}

\vspace{0.5em}
\noindent\textbf{Transformer attention}

In the most recent instantiation of the Transformer attention module~\cite{dai2019transformer}, the attention weight of each query-key pair is computed as the sum of four terms $\{\mathcal{E}_j\}_{j=1}^4$ that are based on different attention factors, as
\begin{equation}
\small
A^{\text{Trans}}_m (q,k,z_q,x_k) \propto  \exp\big(\sum_{j=1}^4 \mathcal{E}_j\big),
\label{eq:attention_relation}
\end{equation}
normalized by $\sum_{k\in \Omega_q}A^{\text{Trans}}_m(q,k,z_q,x_k)=1$ where the supporting key region $\Omega_q$ spans the key elements (\eg, the whole input sentence). By default, $8$ attentional heads are utilized in this paper.

The $\mathcal{E}_1$ and $\mathcal{E}_2$ terms are sensitive to the query content. The $\mathcal{E}_1$ term measures the compatibility of the query and key content, as $\mathcal{E}_1 = z_{q}^\top U_m^\top V^{\rm C}_m x_k$, where $U_m$, $V^{\rm C}_m$ are learnable embedding matrices for the query and key content, respectively. It enables the network to focus more on the keys compatible with the query in terms of content. A possible outcome is the correspondence between similar query and key elements, as illustrated in Fig.~\ref{fig.attention_illustration}~(a). For the $\mathcal{E}_2$ term, it is based on the query content and relative position, as $\mathcal{E}_2 = z_q^\top U_m^\top V^{\rm R}_m R_{k-q}$, where $R_{k-q}$ encodes the relative position $k-q$ by projecting it to a high-dimensional representation through computing sine and cosine functions of different wavelengths\footnote{For 2-d image data, we separately encode the x-axis relative position $R^{\text{X}}_{k-q}$ and y-axis relative position $R^\text{Y}_{k-q}$, and concatenate them to be the final encoding $R_{k-q} = [R^\text{X}_{k-q}, R^\text{Y}_{k-q}]$.} \cite{vaswani2017attention}. $V^{\rm R}_m$ is a learnable embedding matrix for the encoded relative position $R_{k-q}$. This term allows the network to adaptively determine where to assign high attention weights based on the query content. It may help to disentangle appearance from spatial transformations in image recognition, as illustrated in Fig.~\ref{fig.attention_illustration}~(b).

The $\mathcal{E}_3$ and $\mathcal{E}_4$ terms are irrelevant to the query content. The $\mathcal{E}_3$ term involves key content only, as $\mathcal{E}_3 = u_m^\top V^{\rm C}_m {x_k}$, where $u_m$ is a learnable vector. It captures salient key content which should be focused on for the task, and is irrelevant to the query. An illustration is shown in Fig.~\ref{fig.attention_illustration}~(c). As for the $\mathcal{E}_4$ term, it involves relative position only, as $\mathcal{E}_4 = v_m^\top V^{\rm R}_m {R_{k-q}}$, where $v_m$ is a learnable vector. It captures global positional bias between the key and query elements, as illustrated in Fig.~\ref{fig.attention_illustration}~(d).

It is widely believed that query-sensitive prioritization, especially the query and key content compatibility term $\mathcal{E}_1$, is the key to the success of Transformer attention. Thus, in some recent variants~\cite{wang2018non,huang2018ccnet,yuan2018ocnet,fu2018dual}, only $\mathcal{E}_1$ is kept, while the other terms are all removed.

In Transformer attention, both $W_m$ and $W'_m$ in Eq.~\eqref{eq:general_attention} are learnable. $W'_m$ projects the features of $x_k$ to a relatively low dimension for reducing computational overhead, and $W_m$ projects the aggregated features back to the same dimension as $y_q$.

\vspace{0.5em}
\noindent\textbf{Regular and deformable convolution} 

Regular and deformable convolution can be deemed as special instantiations of spatial attention mechanisms, where subsets of the attention factors are involved.

In regular convolution, given a query element, a fixed number of key elements (\eg, $3\times 3$) are sampled, according to predetermined positional offsets with respect to the query. From the perspective of Eq.~\eqref{eq:general_attention}, the attention weight of regular convolution can be expressed as
\begin{equation}
\small
A^{\text{regular}}_m(q, k) =
\begin{cases}
1 & \text{ if $k=q+p_m$} \\
0 & \text{else},
\end{cases}
\end{equation}
where each sampled key element is of a separate attention head (\eg, $3\times 3$ regular convolution corresponds to 9 attention heads), and $p_m$ denotes the offset for the $m$-th sampling position. In addition, the weight $W_m'$ in Eq.~\eqref{eq:general_attention} is fixed as identity, leaving $W_m$ as learnable. In regular convolution, only relative position is involved, without learnable parameters for adapting attention to content. The supporting key region $\Omega_q$ is restricted to a local window centered at the query position and determined by the convolution kernel size.

In deformable convolution~\cite{dai2017deformable,zhu2018deformable}, learnable offsets are added to adjust the sampling positions of the key elements, so as to capture spatial transformations. The learnable offsets are predicted based on the query content, and are thus dynamic to the input. The key and the query elements are from the same set. It can also be incorporated into the generalized attention formulation as a special instantiation of self-attention, where the attention weight is
\begin{equation}
\small
A^{\text{deform}}_m(q, k, x_q) = G(k, q+p_m+w_m^\top x_q),
\end{equation}
where $p_m$ also denotes a predetermined offset, and $w_m^\top x_q$ projects the query content $x_q$ to a deformation offset according to a learnable vector $w_m$\footnote{Following~\cite{dai2017deformable}, the learning rate of $w_m$ is set to 0.1 times that of other parameters to stabilize training.}. $G(a,b)$ is the bilinear interpolation kernel in $N$-d space, which can be decomposed into 1-d bilinear interpolations as $G(a,b) = \prod_{n=1}^{N}g(a_n,b_n)$, where $a_n$ and $b_n$ denote the $n$-th dimension of $a$ and $b$ respectively, and $g(a_n,b_n) = \max (0, 1-|a_n-b_n|)$. Similar to regular convolution, the weight $W_m'$ in Eq.~\eqref{eq:general_attention} is fixed as identity.

In deformable convolution, the attention factors are query content and relative position. The supporting key region $\Omega_q$ can span over all the input elements due to the introduced learnable offsets, while non-zero weights are assigned to a sparse set of key elements where bilinear interpolation is performed. 

\vspace{0.5em}
\noindent\textbf{Dynamic convolution} 

Dynamic convolution~\cite{wu2019pay} is recently proposed to replace the Transformer attention module in self-attention, and is claimed to be simpler and more efficient. It is built upon depth-wise separable convolution~\cite{howard2017mobilenets} with shared dynamic kernel weights, which are predicted based on the query content. In depth-wise separable convolution, a standard convolution is factorized into a depth-wise convolution and a $1\times 1$ convolution called a point-wise convolution, for reducing computation and model size. In depth-wise convolution, a single filter is applied to each input channel, which is fixed for all positions. In dynamic convolution, the kernel weights for the depth-wise convolution are dynamically predicted from the input features, followed by a Softmax normalization. For computational savings, the input channels are divided into several groups, where each group shares the same dynamic kernel weights. In the system of~\cite{wu2019pay}, an orthogonal module called the gated linear unit (GLU)~\cite{dauphin2017language} is applied before the dynamic convolution module to improve accuracy. We include the GLU to respect the original design. 

Dynamic convolution can also be incorporated into the general attention formulation in Eq.~\eqref{eq:general_attention} with minor modifications, where each input feature channel is of a separate attention head. It can be expressed as 
\begin{equation}
\small
y_q = \sum_{c=1}^{C_{\text{in}}}W_c \big[ \sum_{k\in \Omega_q}A^{\text{dynamic}}_c(q,k,x_q) \cdot x_{k,c} \big],
\end{equation}
where $c$ enumerates the channels of the input features ($C_{\text{in}}$ channels in total), $x_{k,c}$ denotes the feature value at the $c$-th channel of $x_k$, and $W_c$ is of the $1\times 1$ point-wise convolution. $A^{\text{dynamic}}_c(q, k, x_q)$ is the attention weight specified by the dynamic kernel in depth-wise convolution, written as
\begin{equation}
\small
A^{\text{dynamic}}_c(q, k, x_q) =
\begin{cases}
K_{j, c} & \text{ if $k=q+p_j$} \\
0 & \text{else},
\end{cases}
\label{eq:attention_dynamic}
\end{equation}
where $p_j$ denotes the $j$-th sampling position in the dynamic kernel, and $K_{j, c}$ is the corresponding kernel weight. Zero attention weight is assigned to keys outside of the kernel. The kernel weight $K_{j, c}$ is predicted from the input features, and is shared among channels in the same group, as
\begin{equation}
\small
K_{j, c} = K_{j, g}^{\text{share}} \propto \exp \big(d_{j,g}^\top x_q \big), \quad  g = \lceil \frac{c}{C_{\text{in}}/N_g} \rceil.
\label{eq:dynamic_weight}
\end{equation}
The input features are divided into $N_g$ groups ($N_g=16$ by default). $K_{j, g}^{\text{share}}$ denotes the dynamic kernel weight for the $g$-th group, and $d_{j,g}$ is the corresponding learnable weight vector. $K_{j, g}^{\text{share}}$ is normalized by $\sum_{j=1}^{N_k} K_{j, g}^{\text{share}} = 1$, where $N_k$ denotes the number of elements in the dynamic kernel.

In dynamic convolution, attention assignment is based on the query content and relative position factor. The supporting key region $\Omega_q$ is restricted to a local window around the query position covered by the dynamic kernel. 

\vspace{0.5em}
\noindent\textbf{Comparing attention mechanisms}

Tab.~\ref{table:compare_attentions} compares the three attention mechanisms discussed above. Transformer attention exploits comprehensive content and position information from both query and key. The $\mathcal{E}_{1}$, $\mathcal{E}_{2}$ and $\mathcal{E}_{4}$ terms require computation proportional to the product of the query and key element numbers, because they involve a traversal of each query-key pair. The $\mathcal{E}_{3}$ term captures key content only, and thus involves computation linear to the key element number. In neural machine translation, the key and query elements are commonly dozens of words in a sentence, so the computational overheads of $\mathcal{E}_{1}$, $\mathcal{E}_{2}$ and $\mathcal{E}_{4}$ are comparable to $\mathcal{E}_{3}$. In image recognition, the key and query elements consist of numerous pixels in an image. The computational overheads of $\mathcal{E}_{1}$, $\mathcal{E}_{2}$ and $\mathcal{E}_{4}$ are thus much heavier than $\mathcal{E}_{3}$.
Note that when the four terms are put together, some computational overhead can be shared among them.

Similar to the $\mathcal{E}_{2}$ term, deformable convolution also is based on query content and relative position. But deformable convolution samples just a sparse set of key elements for each query, and the complexity is linear to the query element number.  Deformable convolution is thus much faster to compute than $\mathcal{E}_{2}$ for image recognition, and is comparable in speed to $\mathcal{E}_{2}$ for machine translation.

Dynamic convolution also relies on query content and relative position. The attention weights of key elements are assigned by the dynamic convolution kernel, based on the query content. Non-zero attention weights only exist in a local range covered by the dynamic kernel. The computational overhead is proportional to the product of the kernel size and query element number. Compared to the $\mathcal{E}_{2}$ term, the computational overhead can be considerably lower if the kernel size is much smaller than the key element number.

We seek to further disentangle the effects of different attention factors, and to facilitate comparison to other instantiations of spatial attention that use a subset of the factors. Thus, manual switches are introduced into the Transformer attention module, which enable us to manually activate / deactivate particular terms. This is expressed as
\begin{equation}
\small
\hat{A}^{\text{Trans}}_m (q,k,z_q,x_k) \propto \exp\big( \sum_{j=1}^4 \beta^{\text{Trans}}_j \mathcal{E}_j \big),
\label{eq:attention_relation_switch}
\end{equation}
where $\{\beta^{\text{Trans}}_j\}$ takes values in $\{0,1\}$ to control the activation of corresponding terms, and $\hat{A}^{\text{Trans}}_m (q,k,z_q,x_k)$ is normalized by $\sum_{k\in \Omega_q} \hat{A}^{\text{Trans}}_m(q,k,z_q,x_k)=1$.

\begin{figure*}
\begin{center}
        \includegraphics[width=0.9\linewidth]{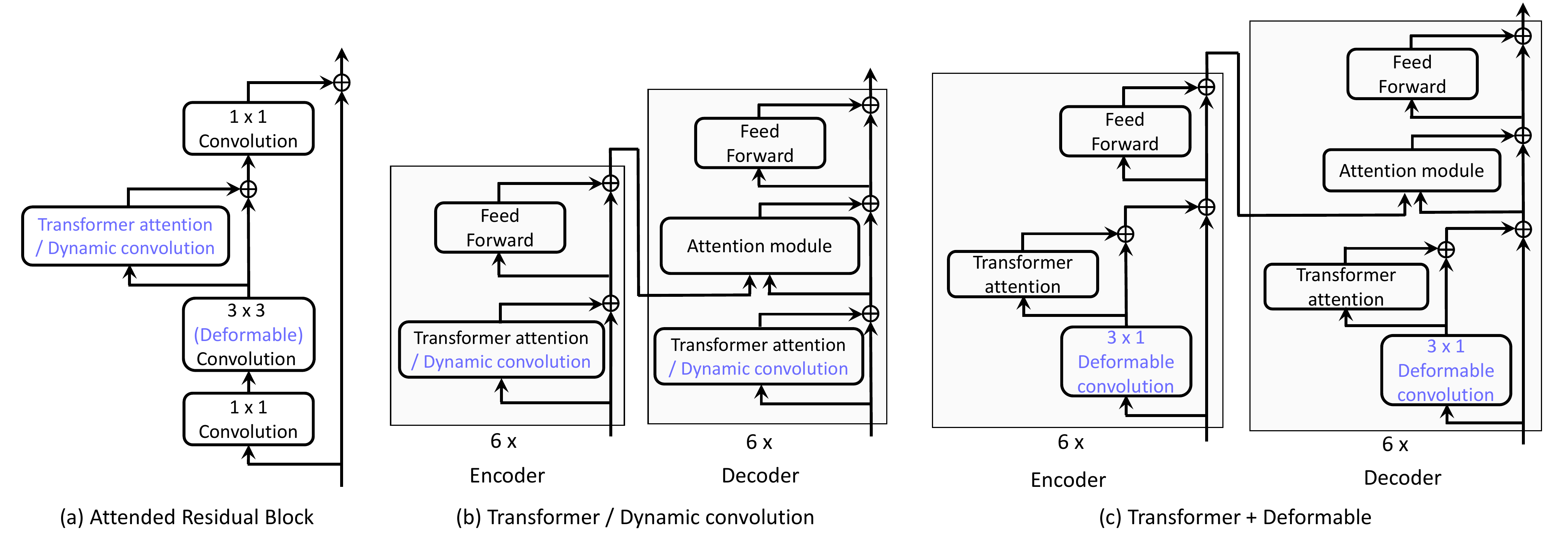}
\end{center}
\vspace{-0.5em}
\caption{Illustration of attention module configurations for empirical study. The modules in blue color are newly added to existing blocks.}
\vspace{-0.8em}
\label{fig.attention_placement}
\end{figure*}

\vspace{0.5em}
\noindent\textbf{Incorporating attention modules into deep networks}

We incorporate various attention mechanisms into deep networks to study their effects.  There are different design choices in inserting the modules, \eg, whether to connect them in series or in parallel, and where to place the modules in the backbone network. We empirically observed the results to be quite similar for different well-considered designs. In this paper, we select the design choices in Fig.~\ref{fig.attention_placement}.
 
For the object detection and semantic segmentation tasks, ResNet-50~\cite{he2016deep} is chosen as the backbone and just the self-attention mechanism is involved. The Transformer attention module is incorporated by applying it on the $3 \times 3$ convolution output in the residual block. For insertion into a pre-trained model without breaking the initial behavior, the Transformer attention module includes a residual connection, and its output is multiplied by a learnable scalar initialized to zero, as in~\cite{wang2018non}. The manner of incorporating dynamic convolution is the same. To exploit deformable convolution, the $3 \times 3$ regular convolution in the residual block is replaced by its deformable counterpart. The resulting architecture is called ``Attended Residual Block", shown in Fig.~\ref{fig.attention_placement} (a). 

In the neuron machine translation (NMT) task, the network architecture follows the Transformer base model~\cite{vaswani2017attention}, where both self-attention and encoder-decoder attention mechanisms are involved. Different from the original paper, we update the absolute position embedding in the Transformer attention module by the latest relative position version~\cite{dai2019transformer} as in Eq.~\ref{eq:attention_relation}. Because both deformable convolution and dynamic convolution capture self-attention, they are added to only the blocks capturing self-attention in Transformer. For dynamic convolution, we replace the Transformer attention module by dynamic convolution directly, as in~\cite{wu2019pay}. The architecture is shown in Fig.~\ref{fig.attention_placement} (b). For its deformable convolution counterpart, because the Transformer model does not utilize any spatial convolution (with kernel size larger than 1), we insert the deformable convolution unit (with kernel size of 3) prior to the input of the Transformer attention module. The resulting architecture is called ``Transformer + Deformable", shown in Fig.~\ref{fig.attention_placement} (c).

\section{Experiments and Analysis}

\subsection{Experimental settings}\label{sec.exp_setting}

\noindent\textbf{Image Object Detection}

Models are trained on the 118k images of the COCO 2017~\cite{lin2014coco} train set. Evaluation is done on the 5k images of the COCO 2017 validation set. Accuracy is measured by the standard mean AP scores at different box IoUs (mAP).

Faster R-CNN~\cite{ren2015faster} with Feature Pyramid Networks (FPN)~\cite{lin2016feature} is chosen as the baseline system. ImageNet~\cite{deng2009imagenet} pre-trained ResNet-50 is utilized as the backbone.  The attended residual blocks in Fig.~\ref{fig.attention_placement} (a) are applied in the last two stages (conv4 and conv5 stages) of ResNet-50. In Transformer attention, the relative position encoding is of the same dimension as the content feature embedding, specifically 256-d and 512-d in the conv4 and conv5 stages, respectively. 

Experiments are implemented based on the open source mmdetection~\cite{mmdetection2018} code base. The hyper-parameter setting strictly follows FPN~\cite{lin2016feature}. Anchors of 5 scales and 3 aspect ratios are utilized. 2k and 1k region proposals are generated at a non-maximum suppression threshold of 0.7 at training and inference respectively. In SGD training, 256 anchor boxes (of positive-negative ratio 1:1) and 512 region proposals (of positive-negative ratio 1:3) are sampled for backpropagating their gradients. In our experiments, the networks are trained on 8 GPUs with 2 images per GPU for 12 epochs. The learning rate is initialized to 0.02 and is divided by 10 at the 8-th and the 11-th epochs. The weight decay and the momentum parameters are set to $10^{-4}$ and 0.9, respectively.

\vspace{0.5em}
\noindent\textbf{Image Semantic Segmentation}

Models are trained on the 5,000 finely annotated images of the Cityscapes~\cite{cordts2016cityscapes} train set. Evaluation is done on the 500 images of the validation set. The standard mean IoU score (mIoU) is used to measure semantic segmentation accuracy.

The CCNet~\cite{huang2018ccnet} for semantic segmentation is utilized, with ImageNet pre-trained ResNet-50 and without the criss-cross attention module proposed in~\cite{huang2018ccnet}, which is a variant of Transformer attention. As done for object detection, the attended residual blocks in Fig.~\ref{fig.attention_placement} (a) are applied in the last two stages.  An additional Transformer attention / dynamic convolution module is placed after the ResNet-50 output following the practice in~\cite{huang2018ccnet} for improving performance. 

The hyper-parameter setting strictly follows that in the CCNet paper~\cite{huang2018ccnet}. In SGD training, the training images are augmented by randomly scaling (from 0.7 to 2.0), randomly cropping (size of 769 $\times$ 769 pixels) and random flipping horizontally. In our experiments, the networks are trained on 8 GPUs with 1 image per GPU for 60k iterations. The ``poly'' learning rate policy is employed, where the initial learning rate is set as 0.005 and multiplied by $ \big( 1 - \frac{\text{iter}}{\text{iter}_{\text{max}}} \big) ^{0.9}$. Synchronized Batch Normalization~\cite{peng2018megdet} is placed after every newly added layer with learnable weights. The weight decay and the momentum parameters are set as $10^{-4}$ and 0.9, respectively.

\vspace{0.5em}
\noindent\textbf{Neural Machine Translation (NMT)}

Model training is conducted on the standard WMT 2014 English-German dataset, consisting of about 4.5 million sentence pairs. Sentences are encoded using byte-pair encoding~\cite{britz2017massive}, with a shared source-target vocabulary of about 37k tokens. Evaluation is on the English-to-German newstest2014 set. Accuracy is measured by the standard bilingual evaluation understudy (BLEU) scores~\cite{papineni2002bleu}.

The Transformer base model~\cite{vaswani2017attention} with relative position encoding~\cite{dai2019transformer} is utilized as the backbone.
Experiments are implemented based on the open source fairseq~\cite{fairseq2017} code base. The hyper-parameters follows the original setting in~\cite{vaswani2017attention}. We used the Adam optimizer~\cite{kingma2014adam} with $\beta_1=0.9$, $\beta_2=0.98$ and $\epsilon=10^{-9}$. In our experiments, the networks are trained on 8 GPUs for 100k iterations. Each training batch contained a set of sentence pairs containing approximately 30k source tokens and 30k target tokens. The initial learning rate is set as $10^{-7}$ and linearly increased to 0.001 after $\text{iter}_{\text{warmup}} = 4000$ iterations, and then multiplied by $\frac{\text{iter}}{\text{iter}_{\text{warmup}}}^{-0.5}$. No weight decay is adopted. During training, label smoothing~\cite{szegedy2016rethinking} of value 0.1 is employed.

\begin{figure*}[t] 
  \centering 
  \subfigure[Image Object detection on COCO]{ 
    \includegraphics[width=0.44\linewidth]{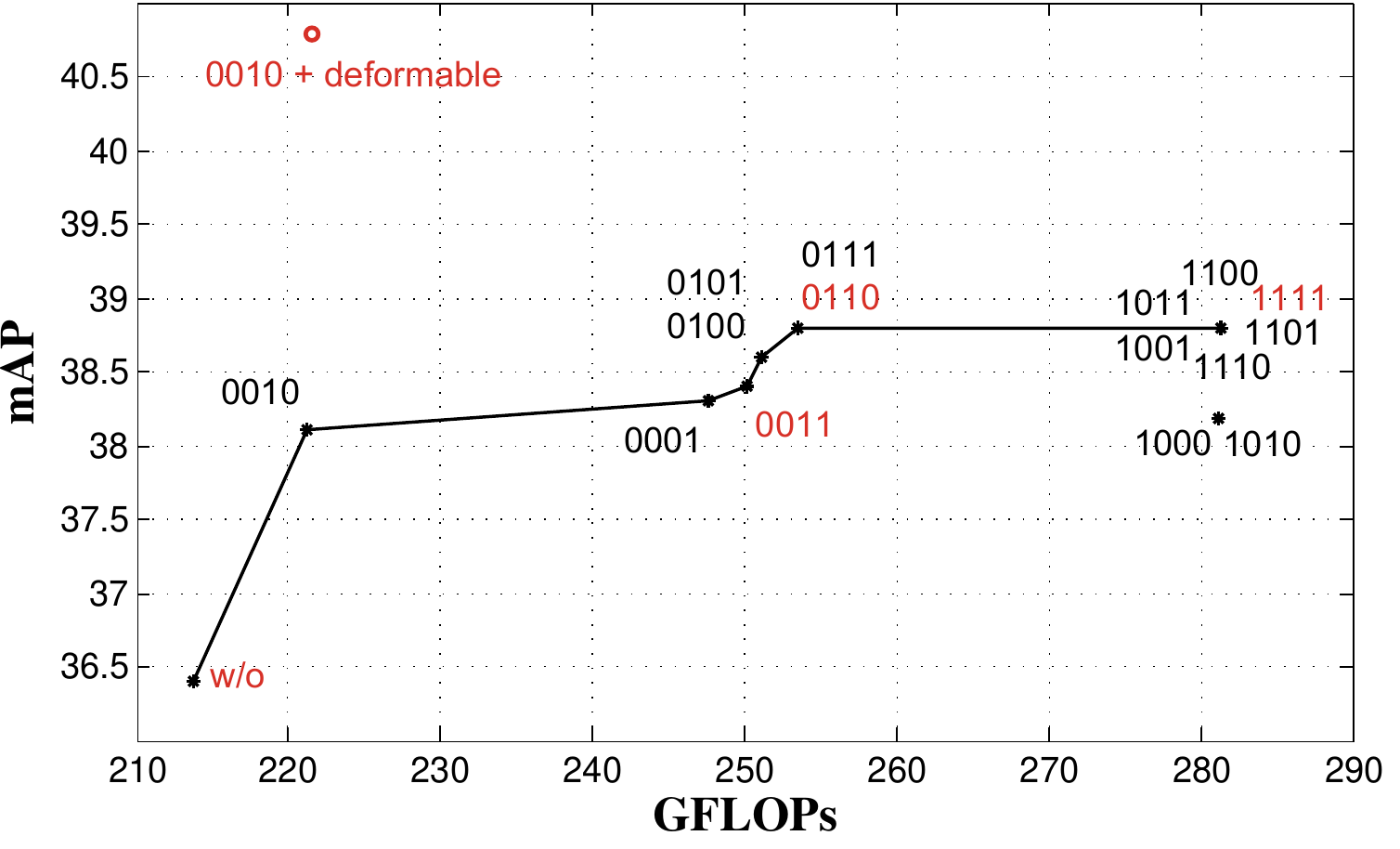}}  
  \vspace{-0.8em}
  \subfigure[Image Semantic Segmentation on Cityscapes]{
    \includegraphics[width=0.44\linewidth]{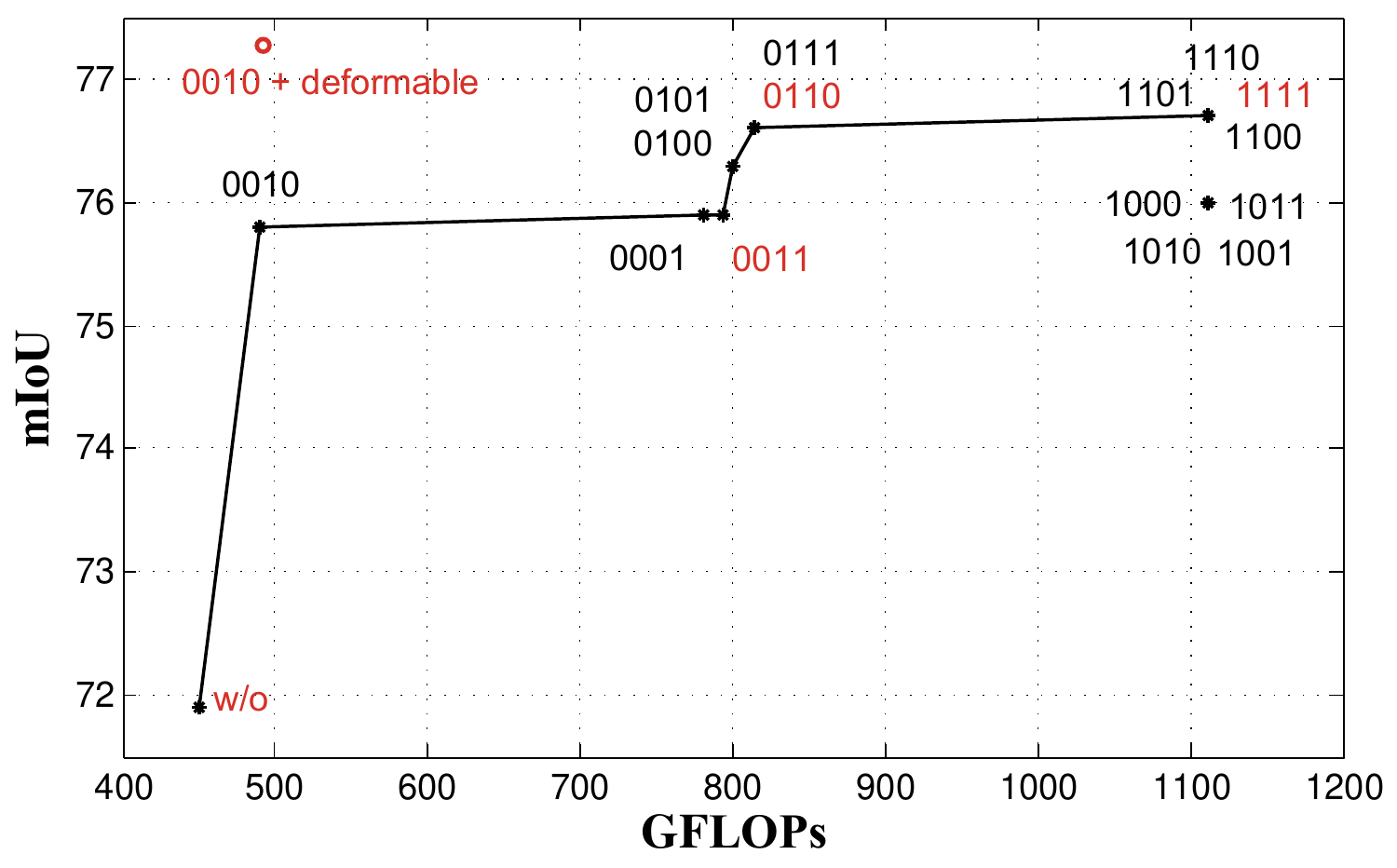}}
  \subfigure[Translation on newstest2014 (self-attention)]{
    \includegraphics[width=0.44\linewidth]{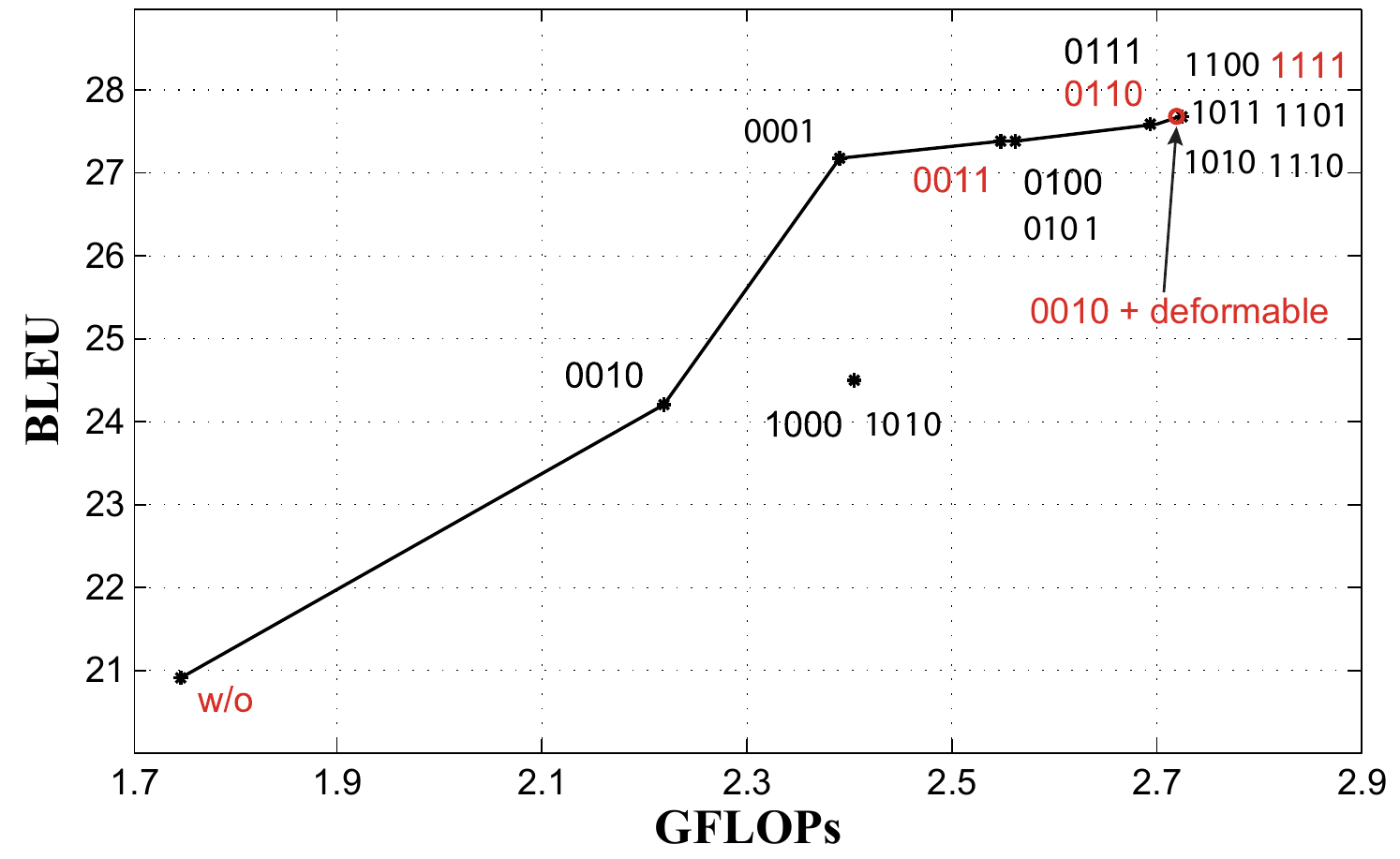}}
  \subfigure[Translation on newstest2014 (encoder-decoder attention)]{
    \includegraphics[width=0.44\linewidth]{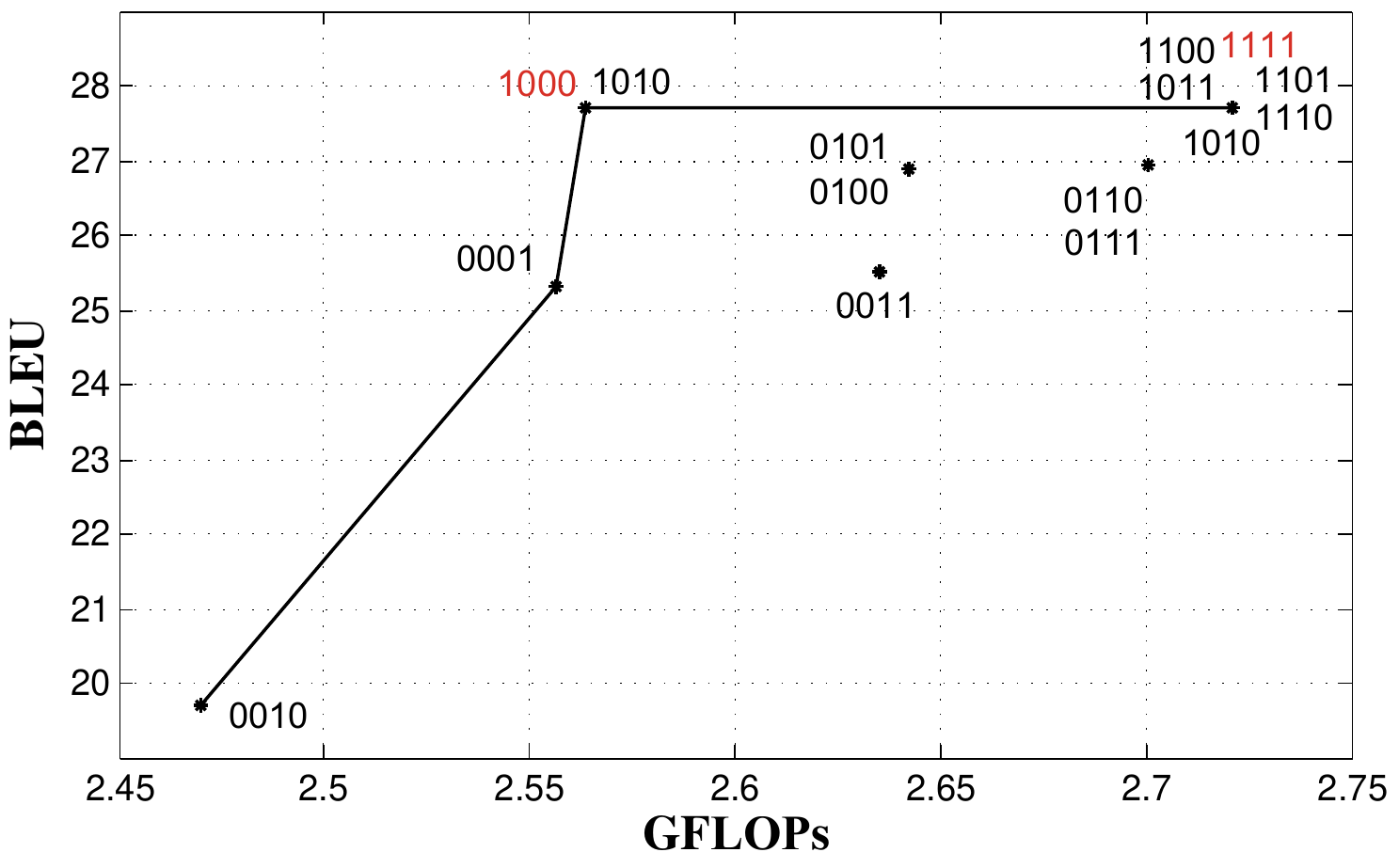}}

  \caption{Accuracy-efficiency tradeoffs of the four terms in Transformer attention ($\mathcal{E}_{1}$ for query and key content, $\mathcal{E}_{2}$ for query content and relative position, $\mathcal{E}_{3}$ for key content only, and $\mathcal{E}_{4}$ for relative position only). The activation and deactivation of particular terms is set by configuration $\{\beta^{\text{Trans}}_{j}\}_{j=1}^{4}$ (\eg, ``0011" denotes the activation of $\mathcal{E}_{3}$ and $\mathcal{E}_{4}$). Because the encoder-decoder attention mechanism is indispensable for NMT, there is no ``w/o'' setting in (d). The results of some configurations overlap in the plots because they are of the same accuracy and computational overhead. The key configurations under study are highlighted in red. The recommended configuration of ``0010 + deformable'' for self-attention in Tab.~\ref{table:ablation_E2_dcn} is also plotted here.}
  \label{fig:decomposing_attention}
  \vspace{-0.3em}
\end{figure*}

\setlength{\tabcolsep}{3pt}
\renewcommand{\arraystretch}{1.1}
\begin{table*}
        \centering
        \small
        \resizebox{1\linewidth}{!}{
        \begin{tabular}{c|c|c|c|c|c|c|c|c|c|c|c|c}
        \Xhline{2\arrayrulewidth}
        \multirow{2}{*}{$\beta^{\text{Trans}}_{1, 2, 3, 4}$ $\rightarrow$ $\beta^{\text{Trans}}_{1, 2, 3, 4}$ + deformable} & \multicolumn{4}{c|}{\makecell{Object Detection (self-attention)}} & \multicolumn{4}{c|}{\makecell{Semantic Segmentation (self-attention)}} & \multicolumn{4}{c}{\makecell{Neural Machine Translation (self-attention)}} \\  
        \cline{2-13}
        & mAP & $\Delta$ mAP & GFLOPs & $\Delta$\% FLOPs & mIoU & $\Delta$ mIoU & GFLOPs & $\Delta$\% FLOPs & BLEU & $\Delta$ BLEU & GFLOPs & $\Delta$\% FLOPs \\
                \hline
                w/o $\rightarrow$ 1111 + deformable & 36.4 $\rightarrow$ \bfseries 41.0 & +4.6 & {\bfseries 213.7} $\rightarrow$ 281.4 & +31.7\% & 71.9 $\rightarrow$ \bfseries 77.8 & +5.9 & {\bfseries 449.5} $\rightarrow$ 1112.1 & +147.4\% & 20.9 $\rightarrow$ \bfseries 28.0 & +7.1 & {\bfseries 1.7} $\rightarrow$ 3.2 & +88.2\%\\  
                \hline
                1111 $\rightarrow$ 1011 + deformable & 38.8 $\rightarrow$ 41.0 & +2.2 & 281.4 $\rightarrow$ 281.4 & -0.0\% & 76.7 $\rightarrow$ 77.8 & +1.1 & 1112.1 $\rightarrow$ 1112.1 & -0.0\% & 27.7 $\rightarrow$ 28.0 & +0.3 & 2.7 $\rightarrow$ 3.2 & +17.3\%\\
                1110 $\rightarrow$ 1010 + deformable & 38.8 $\rightarrow$ 40.9 & +2.1 & 281.4 $\rightarrow$ 281.2 & -0.1\% & 76.7 $\rightarrow$ 77.7 & +1.0 & 1112.1 $\rightarrow$ 1111.2 & -0.1\% & 27.7 $\rightarrow$ 28.0 & +0.3 & 2.7 $\rightarrow$ 2.9 & +5.8\%\\
                1101 $\rightarrow$ 1001 + deformable & 38.8 $\rightarrow$ 41.0 & +2.2 & 281.4 $\rightarrow$ 281.4 & -0.0\% & 76.7 $\rightarrow$ 77.8 & +1.1 & 1112.1 $\rightarrow$ 1112.1 & -0.0\% & 27.7 $\rightarrow$ 28.0 & +0.3 & 2.7 $\rightarrow$ 3.2 & +17.3\%\\
                1100 $\rightarrow$ 1000 + deformable & 38.8 $\rightarrow$ 40.9 & +2.1 & 281.4 $\rightarrow$ 281.2 & -0.1\% & 76.7 $\rightarrow$ 77.7 & +1.0 & 1112.1 $\rightarrow$ 1111.2 & -0.1\% & 27.7 $\rightarrow$ 28.0 & +0.3 & 2.7 $\rightarrow$ 2.9 & +5.8\%\\
                0111 $\rightarrow$ 0011 + deformable & 38.8 $\rightarrow$ 41.0 & +2.2 & 253.6 $\rightarrow$ 250.1 & -1.4\% & 76.6 $\rightarrow$ 77.5 & +0.9 & 814.0 $\rightarrow$ 794.4 & -2.4\% & 27.6 $\rightarrow$ 27.7 & +0.1 & 2.7 $\rightarrow$ 3.0 & +10.9\%\\
                0110 $\rightarrow$ \underline{0010 + deformable} & 38.8 $\rightarrow$ \underline{40.8} & +2.0 & 253.6 $\rightarrow$ \underline{221.1} & -12.8\%& 76.6 $\rightarrow$ \underline{77.3} & +0.7 & 814.0 $\rightarrow$ \underline{489.5} & -39.9\%& 27.6 $\rightarrow$ \underline{27.7} & +0.1 & 2.7 $\rightarrow$ \underline{2.7} & -1.1\%\\
                0101 $\rightarrow$ 0001 + deformable & 38.6 $\rightarrow$ 40.7 & +2.1 & 251.1 $\rightarrow$ 247.6 & -1.4\% & 76.3 $\rightarrow$ 77.3 & +1.0 & 800.7 $\rightarrow$ 781.1 & -2.5\% & 27.4 $\rightarrow$ 27.6 & +0.2 & 2.6 $\rightarrow$ 2.9 & +11.6\%\\
                0100 $\rightarrow$ w/o + deformable     & 38.6 $\rightarrow$ 39.9 & +1.3 & 251.1 $\rightarrow$ 213.7 & -14.9\%& 76.3 $\rightarrow$ 77.2 & +0.9 & 800.7 $\rightarrow$ 449.5 & -43.9\%& 27.4 $\rightarrow$ 27.3 & -0.1 & 2.6 $\rightarrow$ 2.2 & -13.5\%\\
        \Xhline{2\arrayrulewidth}
        \end{tabular}
        }
        \vspace{.5em}
        \caption{Deformable convolution vs. $\mathcal{E}_2$ in Transformer attention, where both exploit query content and relative position information. The underlined configuration of ``0010 + deformable'' is recommended for an optimal accuracy-efficiency tradeoff.}
        \label{table:ablation_E2_dcn}
        \vspace{-0.5em}
\end{table*}

\setlength{\tabcolsep}{3pt}
\renewcommand{\arraystretch}{1.1}
\begin{table*}
        \centering
        \small
        \resizebox{1\linewidth}{!}{
        \begin{tabular}{c|c|c|c|c|c|c|c|c|c|c|c|c}
        \Xhline{2\arrayrulewidth}
        \multirow{2}{*}{$\beta^{\text{Trans}}_{1, 2, 3, 4}$ $\rightarrow$ dynamic} & \multicolumn{4}{c|}{\makecell{Object Detection (self-attention)}} & \multicolumn{4}{c|}{\makecell{Semantic Segmentation (self-attention)}} & \multicolumn{4}{c}{\makecell{Neural Machine Translation (self-attention)}} \\  
        \cline{2-13}
        & mAP & $\Delta$ mAP & GFLOPs & $\Delta$\% FLOPs & mIoU & $\Delta$ mIoU & GFLOPs & $\Delta$\% FLOPs & BLEU & $\Delta$ BLEU & GFLOPs & $\Delta$\% FLOPs \\
                \hline
                \bfseries 0100 & \bfseries 38.6 & - & \bfseries 251.1 & - & \bfseries 76.3 & - & \bfseries 800.7 & - & \bfseries 27.4 & - & \bfseries 2.6 & - \\
                \hline
                0100 ($n_k=31$) $\rightarrow$ dynamic ($n_k=31$) & 38.6 $\rightarrow$ 37.9 & -0.7 & 229.4 $\rightarrow$ 352.9 & +53.8\% & 75.5 $\rightarrow$ 74.2 & -1.3 & 523.3 $\rightarrow$ 1029.0 &+96.6\% & 27.4 $\rightarrow$ 27.6 & +0.2 & 2.4 $\rightarrow$ 2.4 & +1.8\% \\
                0100 ($n_k=25$) $\rightarrow$ dynamic ($n_k=25$) & 38.6 $\rightarrow$ 37.8 & -0.8 & 226.6 $\rightarrow$ 306.8 & +35.4\% & 75.5 $\rightarrow$ 74.2 & -1.3 & 511.8 $\rightarrow$ 840.4 & +64.2\% & 27.4 $\rightarrow$ 27.6 & +0.2 & 2.3 $\rightarrow$ 2.3 & +1.4\% \\
                0100 ($n_k=19$) $\rightarrow$ dynamic ($n_k=19$) & 38.6 $\rightarrow$ 37.6 & -1.0 & 224.4 $\rightarrow$ 270.6 & +20.6\% & 75.4 $\rightarrow$ 73.7 & -1.7 & 502.6 $\rightarrow$ 692.1 & +37.7\% & 27.4 $\rightarrow$ 27.5 & +0.1 & 2.3 $\rightarrow$ 2.3 & +1.1\% \\
                0100 ($n_k=13$) $\rightarrow$ dynamic ($n_k=13$) & 38.5 $\rightarrow$ 37.5 & -1.0 & 222.7 $\rightarrow$ 244.3 & +9.7\%  & 74.4 $\rightarrow$ 71.9 & -2.5 & 495.9 $\rightarrow$ 584.3 & +17.8\% & 27.3 $\rightarrow$ 27.4 & +0.1 & 2.3 $\rightarrow$ 2.3 & +0.7\% \\
        \Xhline{2\arrayrulewidth}
        \end{tabular}
        }
        \vspace{.5em}
        \caption{Dynamic convolution vs. $\mathcal{E}_2$ in Transformer attention, where both exploit query content and relative position information.  The kernel size of dynamic convolution $N_k$ is $n_k^2$ for image recognition and $n_k$ for NMT. The spatial range of Transformer attention is also constrained to be the kernel size of dynamic convolution for ablation.}
        \label{table:ablation_E2_dynamic}
        \vspace{-1em}
\end{table*}

\subsection{Effects of various attention-based modules}

\noindent\textbf{Disentanglement in Transformer attention}

We first seek to disentangle the effects of the four terms in the Transformer attention module. This is achieved by manually setting the $\{\beta^{\text{Trans}}_{j}\}_{j=1}^{4}$ values in Eq.~\eqref{eq:attention_relation_switch} to control the activation / deactivation of individual terms. The network is trained and tested for all 16 possible configurations of $\{\beta^{\text{Trans}}_{j}\}_{j=1}^{4}$. In this set of experiments, no other attention mechanisms are involved. Thus, for the object detection and semantic segmentation tasks, the $3\times 3$ convolution is of regular convolution in the network of Fig.~\ref{fig.attention_placement} (a). For the NMT task, the network architecture in Fig.~\ref{fig.attention_placement} (b) is utilized. Transformer attention is used in the choices of ``Transformer attention / Dynamic convolution" in Fig.~\ref{fig.attention_placement} (a) and (b). Note that for the NMT task, Transformer attention modules are utilized for both self-attention and encoder-decoder attention. To reduce experimental complexity, the Transformer attention modules in encoder-decoder attention are kept as their full version ($\beta^{\text{Trans}}_{j}=1, j=1,\ldots,4$, abbreviated as configuration ``1111" here) when we study self-attention.

Fig.~\ref{fig:decomposing_attention} plots the accuracy-efficiency tradeoffs of different $\{\beta^{\text{Trans}}_{j}\}_{j=1}^{4}$ configurations, where the accuracy-efficiency envelopes are indicated by connected line segments. Note that only the computational overheads from the Transformer attention modules under study are counted here, without the overheads from other parts of the network. From the plot, we draw the following conclusions:

(1) \textit{In self-attention, the query-sensitive terms play a minor role compared to the query-irrelevant terms.} Especially, the query and key content term have a negligible effect on accuracy, while being computationally heavy in image recognition tasks. Overall, the accuracy gain brought by the Transformer attention module is large (from the configuration where the Transformer attention module is removed (``w/o") to that where the full version of Transformer attention is utilized (``1111")). It can be seen that the gain brought by the query-irrelevant terms (from configuration ``w/o" to ``0011") is much larger than that brought by the query-sensitive terms (from configuration ``0011" to ``1111"). Particularly, the performance gain brought by the query and key content term (controlled by $\beta^{\text{Trans}}_{1}$) is negligible. Removing it (from configuration ``1111" to ``0111") incurs only a tiny drop in accuracy, while considerably reducing the computational overhead in image recognition tasks.

(2) \textit{In encoder-decoder attention, the query and key content term is vital.} Deactivation of it (controlled by $\beta^{\text{Trans}}_{1}$) incurs a noticeable drop in accuracy, while only utilizing the query and key content term (configuration ``1000") delivers accuracy almost the same as the full version (configuration ``1111"). This is because the key step in NMT is to align the words in the source and the target sentences. A traversal of the query and key content is essential for such alignment.

(3) \textit{In self-attention, the attention factors of query content \& relative position and the key content only are most important.} The corresponding configuration ``0110" delivers accuracy very close to the full version (configuration ``1111"), while saving a considerable amount of computational overhead in image recognition tasks. It is also worth noting that the key content only term, which captures saliency information, can effectively improve the performance with little additional overhead.

Our findings contradict the widespread belief that query-sensitive terms, especially the query and key content term, are crucial for the success of Transformer attention. The experimental results suggest that this is only true for the encoder-decoder attention scenario. In self-attention scenarios, the query and key content term is even removable.

\vspace{0.5em}
\noindent\textbf{Deformable convolution vs. $\mathcal{E}_2$ in Transformer attention}

Here, we compare deformable convolution and the $\mathcal{E}_{2}$ term from Transformer attention in Eq.~\eqref{eq:attention_relation}. Because deformable convolution is designed for capturing self-attention, we restrict the experiments to self-attention scenarios only. Note that when deformable convolution is utilized in the NMT task, the network architecture is of ``Transformer + Deformable" in Fig.~\ref{fig.attention_placement} (c). 

Tab.~\ref{table:ablation_E2_dcn} compares deformable convolution and the $\mathcal{E}_{2}$ term in a variety of settings. We find that:

(1) \textit{For object detection and semantic segmentation, deformable convolution considerably surpasses the $\mathcal{E}_{2}$ term in both accuracy and efficiency.} While for NMT, deformable convolution is on par with the $\mathcal{E}_{2}$ term in both accuracy and efficiency. In terms of efficiency, deformable convolution does not need to traverse all the key elements. This advantage is obvious on images, where numerous pixels are involved. In terms of accuracy, the bilinear sampling in deformable convolution is based on the hypothesis of local linearity of feature maps. This hypothesis holds better on images where local image content changes gradually, than on languages where words change abruptly.  

(2) \textit{The combination of deformable convolution and the key content only term (``0010 + deformable") delivers the best accuracy-efficiency tradeoff.} The accuracy is on par with using deformable convolution and the whole attention module (``1111 + deformable''), while the overhead is slightly higher than that of deformable convolution only (``w/o\ + deformable''). This finding is in line with finding (3) of ``Disentanglement in Transformer attention". It further suggests the importance of the query content \& relative position and key content only factors in self-attention. The configuration ``0010 + deformable" is also plotted in Fig.~\ref{fig:decomposing_attention}.

\vspace{0.5em}
\noindent\textbf{Dynamic convolution vs. $\mathcal{E}_2$ in Transformer attention}

We compare these two instantiations in self-attention scenarios. The network architectures are of  Fig.~\ref{fig.attention_placement} (a) for image recognition tasks, and of Fig.~\ref{fig.attention_placement} (b) for NMT, where either the Transformer attention with $\mathcal{E}_2$ only (configuration ``0100") or dynamic convolution is utilized.

Tab.~\ref{table:ablation_E2_dynamic} presents the results. We can find that for NMT, dynamic convolution achieves accuracy on par with the $\mathcal{E}_{2}$ term at reduced computational cost. However, dynamic convolution is not effective for object detection and semantic segmentation, delivering considerably lower accuracy. To further study the influence of kernel size in dynamic convolution, we also constrain the spatial range of the $\mathcal{E}_2$ term to be the same as that in dynamic convolution. The accuracy drops as the spatial range shrinks for both dynamic convolution and the $\mathcal{E}_{2}$ term. But it is worth noting that the $\mathcal{E}_{2}$ term still surpasses dynamic convolution at the same spatial range in image recognition tasks, with even smaller computational overhead. The inferior accuracy of dynamic convolution in image recognition tasks might be because dynamic convolution is originally designed for NMT, and some design choices may not be suitable for image recognition.

{\small
\bibliographystyle{ieee}
\bibliography{egbib}
}

\end{document}